\documentclass[10pt,twocolumn,letterpaper]{article}

\usepackage{iccv}
\usepackage{times}
\usepackage{epsfig}
\usepackage{graphicx}
\usepackage{amsmath}
\usepackage{amssymb}

\usepackage{bm}
\usepackage{multirow}
\usepackage{booktabs}

\usepackage[pagebackref=true,breaklinks=true,letterpaper=true,colorlinks,bookmarks=false]{hyperref}

\iccvfinalcopy

\begin{document}

\title{Dual Normalization Multitasking for Audio-Visual Sounding Object Localization}

\author{
Tokuhiro Nishikawa, Daiki Shimada, Jerry Jun Yokono\\
Sony Group Corporation, Tokyo, Japan\\
{\tt\small \{tokuhiro.nishikawa, daiki.shimada, jerry.jun.yokono\}@sony.com}}

\maketitle

\begin{abstract}
   Although several research works have been reported on audio-visual sound source localization in unconstrained videos, no datasets and metrics have been proposed in the literature to quantitatively evaluate its performance.
   Defining the ground truth for sound source localization is difficult, because the location where the sound is produced is not limited to the range of the source object, but the vibrations propagate and spread through the surrounding objects.
   Therefore we propose a new concept, Sounding Object, to reduce the ambiguity of the visual location of sound, making it possible to annotate the location of the wide range of sound sources.
   With newly proposed metrics for quantitative evaluation, we formulate the problem of Audio-Visual Sounding Object Localization (AVSOL).
   We also created the evaluation dataset (AVSOL-E dataset) by manually annotating the test set of well-known Audio-Visual Event (AVE) dataset \cite{tian2018audio}.
   To tackle this new AVSOL problem, we propose a novel multitask training strategy and architecture called Dual Normalization Multitasking (DNM), which aggregates the Audio-Visual Correspondence (AVC) task and the classification task for video events into a single audio-visual similarity map.
   By efficiently utilize both supervisions by DNM, our proposed architecture significantly outperforms the baseline methods.
\end{abstract}

\section{Introduction}
\label{sec:intro}
When we hear a dog barking, we associate the sound of the bark with the appearance of the dog, and naturally perceive it as a single event.
A certain visual input comes in conjunction with a certain auditory input, and we can relate both information to localize the source object.
By properly processing visual and audio multimodal inputs complementary, we better understand and explore the world around us.

If we could realize the same ability on machines, behaviors of the real-world robots will become more intelligent.
For example, a robot can pay attention to the person who is actually talking to it among multiple people,
or a robot and a person can pay joint attention to a specific object that is making a sound, which gives the person an opportunity to teach the robot the name of the object.

\begin{figure}[t]
   \begin{center}
      \includegraphics[width=0.48\linewidth]{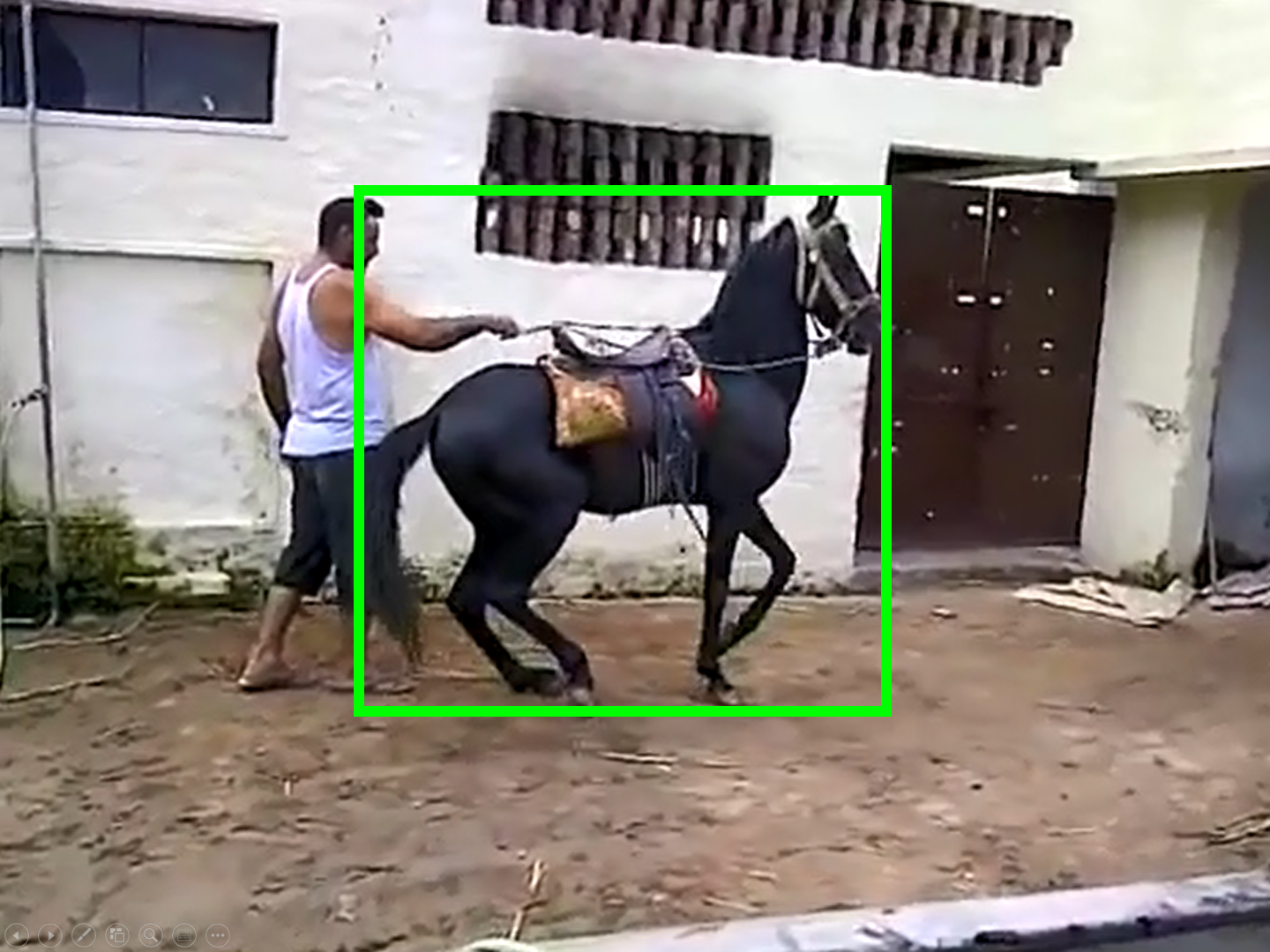}
      \includegraphics[width=0.48\linewidth]{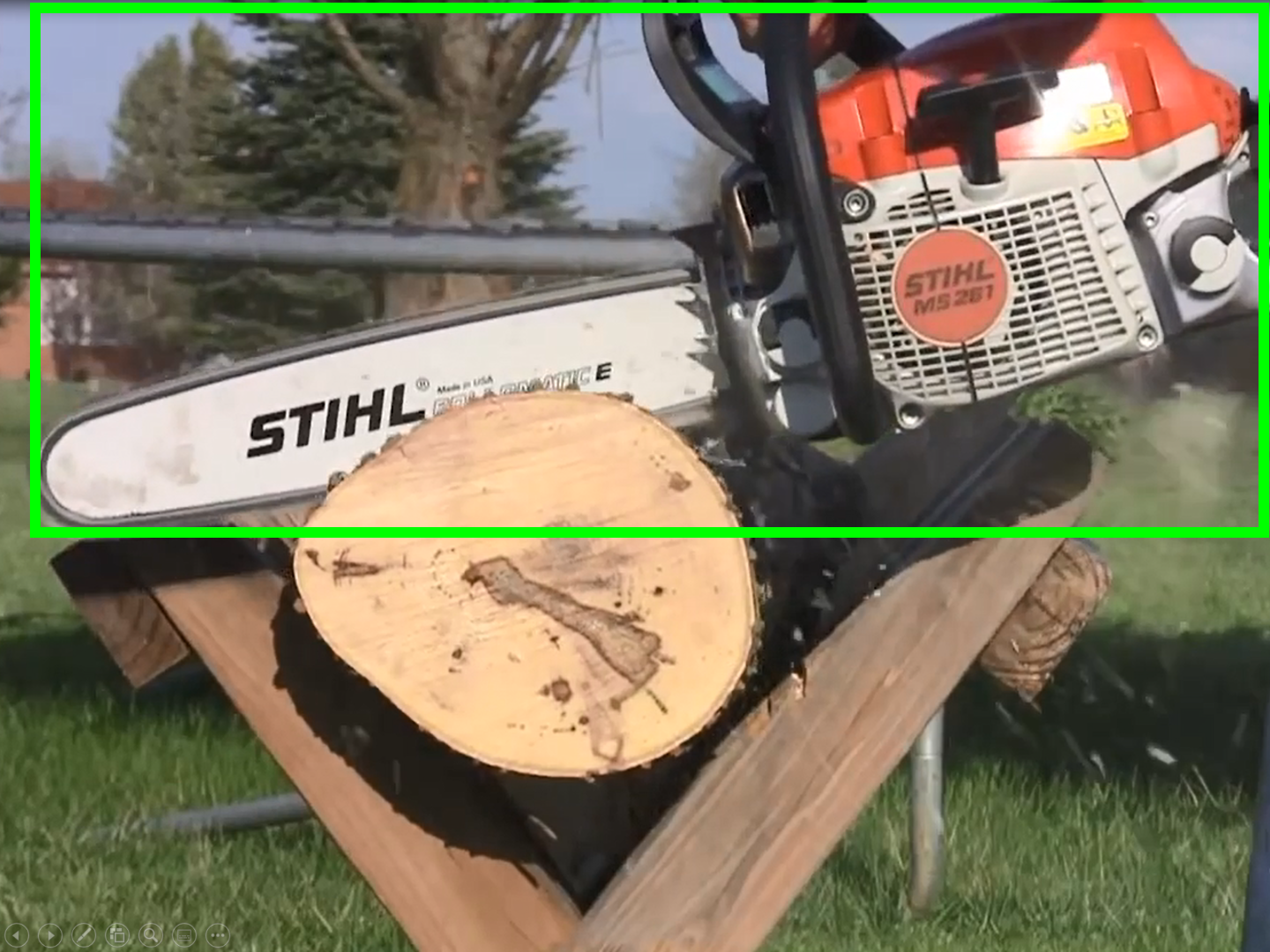}
   \end{center}
   \caption{
      Examples of {\textbf {\textit{Sounding Object}}}.
      The visual location of sound source is inherently ambiguous.
      The sound of a horse walking is actually the sound of its hooves hitting the ground (left).
      The sound of a chainsaw is produced by the engine, moving chain, vibration of the log being cut, flying wood chips, \etc (right).
      We introduce the concept of {\textit{Sounding Object}} to avoid this ambiguity, and make it possible to annotate their locations in the similar way to regular object detection (green bounding-box).
   }
   \label{fig:soundingobject}
\end{figure}

There have been a lot of research in the field of audio-visual sound source localization, \ie a technique for visually locating a sound source.
One of the biggest problems in the field is there is no method to quantitatively evaluate the performance of localization in unconstrained videos.
Therefore, each study can only qualitatively evaluate the performance by visualization.
Senocak \etal\cite{senocak2018learning} proposed to create annotations with the weighted ground truth of the sound location by the consensus of multiple annotators.
However, it is often seen that the opinions of each annotators differ, due to the ambiguity of the location of sound sources.
Since sound is generated by vibrations of materials, it is not possible to specify exactly where the sound is produced.

We tackle this problem with the following ideas:
(1) locate the object which is the original cause of the sound;
(2) target the whole object, not a part of it;
(3) exclude ambient sounds and limit our target to object sounds.
As we describe in Section \ref{sec:soundingobject}, these ideas eliminate much of the ambiguity about determining the location of the sound.
We call such an object a {\it Sounding Object} (Figure \ref{fig:soundingobject}).

Our goal is to localize the sounding object in unconstrained videos.
We call this problem the {\it Audio-Visual Sounding Object Localization} (AVSOL) and create the AVSOL evaluation (AVSOL-E) dataset for quantitative evaluation.
We manually annotated ground truth of sounding objects by bounding-boxes for the videos in the test set of the AVE dataset which is a widely used dataset created by Tian \etal\cite{tian2018audio}
We also propose three new evaluation metrics for AVSOL: HmBoxAUC, PiBR and PNSR.
Details are described in Section \ref{sec:avsole}.
Since the definition of ground truth is similar to that of visual-only object localization or detection, it has high affinity with industrial applications.
The AVSOL-E dataset will be publicly available on GitHub\footnote{https://github.com/sony/ai-research-code}.

To solve AVSOL, we propose a novel algorithm with a double-head architecture, called Dual Normalization Multitasking (DNM).
DNM connects the one audio-visual similarity map (AVSM) with two different tasks (AVC and event classification) simultaneously to efficiently utilize the information obtained from these two tasks.

The contributions of this research are twofold:
(1) For the first time in this field, we create AVSOL-E dataset and evaluation metrics, to make it possible to quantitatively evaluate the performance of AVSOL methods;
(2) We propose a novel DNM architecture for AVSOL which significantly outperforms baseline methods.

\section{Related Work}
\noindent
{\bf Sound Source Localization.}
Sound source localization by deep models is well studied in recent years.
Deep learning requires a large labeled dataset, but it is difficult to prepare labels indicating the sound source locations in such a scale.
Therefore, the mainstream is based on self- or weakly-supervised learning, which achieves localization without any supervisory data indicating the location of sound.

In early studies, the activation of a visual network trained with audio as supervisory signals is used to estimate the location\cite{owens2016ambient, arandjelovic2017look}.
Arandjelovi\'c and Zisserman \cite{arandjelovic2017look} introduced AVC task, \ie binary classification of whether the sound and video match or not.
However, since these methods do not use sound input during inference, they only indicate the location of objects that are likely to produce sound.

Many of the recent techniques\cite{owens2018audio, arandjelovic2018objects,senocak2018learning,tian2018audio,ramaswamy2020see,oya2020we} use MIL, Class Activation Mapping (CAM) or attention to localize the sound source.
In \cite{owens2018audio}, an early fusion network of sound and video is introduced and CAM is used to localize the sound source.
In \cite{arandjelovic2018objects}, a late fusion network which compares sound and pixel-wise visual features is proposed and trained by the AVC task.
In their architecture, each pixel of the visual feature corresponds to a single instance of MIL, and succeeded to recognize the rough shape of the sound source object.
In \cite{senocak2018learning} audio-visual attention map which indicates the sound source is calculated and applied to visual feature to solve AVC task.
Although these methods successfully combined the AVC task with MIL or attention, they used still images, not video, as visual input.

Tian \etal\cite{tian2018audio} created AVE dataset and used event classification for training.
Their system learns to align the attention to the audio-visual event.
Although they also proposed an audio-visual distance learning similar to the AVC, they did not combine it with event classification.
We show the localization performance is further boosted by multitasking AVC and event classification with our DNM architecture.

Dealing with multiple sound sources is a difficult issue.
Methods have been proposed for localizing multiple sound sources\cite{hu2019deep,hu2020discriminative,qian2020multiple}.
In \cite{qian2020multiple}, multitask learning of AVC and event classification is used in the two-stage learning framework and a fine-grained audio-visual alignment is performed.
We also leverage the supervision from these two tasks but in much simpler way.
In spite of its simplicity, we show localizing multiple sound sources is possible.

\noindent
{\bf Evaluating Sound Source Localization.}
As described in Section \ref{sec:intro}, \cite{senocak2018learning} proposed a consensus based annotation and collected the sound source localization dataset.
Their dataset is based on Flickr-SoundNet\cite{aytar2016soundnet} which consists of sound and still images.
Recently the dataset of \cite{senocak2018learning} is reported to be problematic.
\cite{oya2020we}
showed the sound source in the dataset can be estimated using only visual keys.
Hu \etal\cite{hu2020discriminative} used Faster RCNN\cite{ren2015faster} trained to detect instruments, to generate ground truth for a dataset of music play (MUSIC\cite{zhao2018sound}, AudiosSet-Instrument\cite{gao2019co}).
Since instruments are objects which clearly designed for producing sound, it is easy to use the detector to make ground truth.

\noindent
{\bf Weakly Supervised Object Localization.}
Weakly supervised object localization (WSOL)
is commonly set up as a problem to extract object regions in the image given only a supervised signal of its category.
Traditionally WSOL have been studied using MIL\cite{Carbonneau_2018} and is improved by introducing prior knowledge such as symmetry\cite{bilen2014weakly} or motion\cite{prest2012learning}, or by addressing the problem of local optimization\cite{gokberk2014multi}.

Recently, CAM\cite{zhou2016learning} is often used to extract objects location at pixel level.
CAM is shown to be a kind of MIL that instantiates a receptive field corresponding to each pixel on image features\cite{choe2020evaluating}.
Attention\cite{jetley2018learn} is also shown to be an another form of MIL\cite{ilse2018attention}.
CAM and attention, as WSOL, has been criticized for highlighting mainly discriminative regions, and technics such as data augmentations and architectural solutions have been proposed to extract the entire object\cite{singh2017hide,yun2019cutmix,zhang2018adversarial,choe2019attention}.
However, recent study\cite{choe2020evaluating} reported a series of CAM based methods rely on implicit full supervision for setting hyperparameter (\eg score map threshold $\tau$), and did not reach CAM when evaluated by fair protocol.

Based on these findings, we propose evaluation metrics that is not affected by the threshold $\tau$ for the heatmap.
Our DNM aggregates MIL for AVC and attention for event classification into a single AVSM, and make it possible to train our model by these two tasks without inconsistencies.

\section{Audio-Visual Sounding Object Localization}
\label{sec:avsol}

\subsection{Sounding Object}
\label{sec:soundingobject}
We define the sounding object as described in Section \ref{sec:intro}.
For example, in Figure \ref{fig:soundingobject}, the sound of a horse running is produced not by the horse itself, but by the collision of the horse's hoof with the ground.
Therefore, in a physical sense, the sound source localization should output a certain range that includes the horse's hoof and the ground.
Since the physical vibration is spread out through materials, it is never possible to spatially define a precise answer to the location of sound source.
In this situation, however, if we ask the original cause of the sound, then it will be a "horse running", and the bounding-box annotation should be given to the horse by the definition of sounding object.

In a scene where a chainsaw is used to cut logs, the sound of the engine, the blade rotating
and the vibration of the wood being cut, are all mixed together, making it difficult to determine the actual physical source of the sound.
However, if we follow the definition of sounding object, estimating only the region of the chainsaw would be a desirable.
Note that this "object" oriented annotations are very important for many real-world applications such as robot interactions and object-level image retrievals.

\subsection{Dataset Description}
With the definition of sounding object, we can now annotate a wide range of video.
We create annotations for the videos in the test set of unconstrained AVE dataset\cite{tian2018audio}.
AVE denotes an event which is both visible and audible in the scene.
The AVE dataset is a subset of Audioset\cite{gemmeke2017audio}, and contains 4143 unconstrained videos each of 10 seconds, across 28 different categories which cover a wide range of AVEs from different domains \eg humans, animals, vehicles, musics and machines.
All categories of events are with object sounds, and there is no ambient sound event.
Each video contains at least one 2 seconds long AVE, \ie it may contains non-AVE segments.

\begin{table}
   \begin{center}
      \begin{tabular}{llrr}
         \toprule
         \multirow{2}{*}{AVE}     & single object    & 24,005 & \multirow{2}{*}{30,791} \\
                                  & multiple objects & 6,786  &                         \\
         \midrule
         \multirow{3}{*}{Non-AVE} & visible          & 5,183  & \multirow{3}{*}{6,509}  \\
                                  & audible          & 860    &                         \\
                                  & neither (noise)  & 366    &                         \\
         \bottomrule
         \addlinespace
         \multicolumn{4}{r}{Total: $373~{\rm videos} \times 10~{\rm sec.} \times 10~{\rm fps} = 37,300$}
      \end{tabular}
   \end{center}
   \caption{\# of annotated video frames in AVSOL-E dataset.}
   \label{table:statistics}
\end{table}

\noindent
{\bf The AVSOL-E dataset.}
For each 10 seconds video in the test set of AVE dataset, we created bounding-box annotations for sounding objects at 10 fps.
The test set contains 403 videos but only 373 videos were annotated for AVSOL-E because we excluded irregular videos (\eg Game CGs, cartoons, videos whose audio is edited in post-processing).
Each video may show multiple sounding objects at the same time in a single frame.
In scenes like those, we annotated each sounding object, regardless of whether they are of the same category or a different category.

The Non-AVE frames may contain the event which is only visible (\ie potential sounding object which is not making sound) or only audible (\ie out-of-view sound), or neither (\ie noise).
bounding-boxes are given as long as the target is visible, regardless of whether or not the target is making a sound.
Only when the visible object is making a sound at the moment, a 'sounding' tag is given to the target's bounding-box.
To identify the only audible part, we introduce a 'out-of-view' tag.
When the sound is audible and there is no visible object corresponds to it, we create a dummy bounding-box and attach the 'out-of-view' tag.
Table \ref{table:statistics} shows the statistics of AVSOL-E dataset.

\subsection{Evaluation Metrics}
\label{sec:avsole}

We evaluate performance of AVSOL in three perspectives.
In the frames of the AVE segments, we evaluate (1) whether the heatmap is spread out in the target area, and (2) whether the location of the heatmap peak is correct.
In non-AVE frames, as it's desirable not to localize target, we evaluate (3) how well the output heatmap is suppressed.

\noindent
{\bf HmBoxAUC.}
Heatmap vs Bounding-Box Area Under the Curve (HmBoxAUC) is introduced to evaluate localization performance in AVE frames.
For a single frame,
let the index of each pixel be $i$.
Each pixel value $g_i$ of ground truth is either $g_i=1$ for foreground (\ie the pixel is in the bounding-box of sounding object) or $g_i=0$ for background.
The each pixel value of heatmap $h_i$ is binarized by the threshold $\tau$.
We define precision and recall between heatmap and bounding-box ground truth as follows.
\begin{eqnarray}
   {\rm precision}(\tau) & = & \frac{|\{h_i \geq \tau\} \cap \{g_i = 1\}|}{|\{h_i \geq \tau\}|}\\
   {\rm recall}(\tau) & = & \frac{|\{h_i \geq \tau\} \cap \{g_i = 1\}|}{|\{g_i=1\}|}
\end{eqnarray}

For threshold independence, we use
$ {\rm HmBoxAUC} = \sum_d {\rm precision}(\tau_d)({\rm recall}(\tau_d)-{\rm recall}(\tau_{d-1})) $.

\noindent
{\bf PiBR.}
Peak in Box Ratio (PiBR) is to quantify the pinpoint localization performance.
It is simply the percentage of the number of frames where the peak of the heatmap is within the ground truth bounding-boxes, for all evaluation frames in the AVE segments.
If there are multiple sounding objects, it is counted as correct if the peak is in one of them.

\noindent
{\bf PNSR.}
Since our model learn to generate localization map of the sounding object by the AVC task, the discrimination between AVE and non-AVE frames can be predicted by the output level of the map.
In AVE frames, the peak of the map should be on the sounding objects and the value should be high.
Conversely in non-AVE frames, the peak should be suppressed.
We introduce Peak Noise to Signal Ratio (PNSR) to measure this performance.

For a single image frame, let the number of sound sources be $L$, and the bounding-boxes of sounding object be $B = \{B_1, B_2, ...,B_L\}$.
We assign the number $j=1,2,...,J$ to all the evaluation frames in the test videos.
Let $F_{\rm AVE}$ be the set that contains all AVE frames, then the PNSR is as follows.
\begin{equation}
   {\rm PNSR} = \frac{{\rm average}_{j \notin F_{\rm AVE}} \max_i( h_i^j )} {{\rm average}_{j \in F_{\rm AVE}} \max_{i \in B^j}( h_i^j )}
\end{equation}
Unlike the general signal-to-noise ratio, the closer the PNSR is to zero, the better the performance.

\begin{figure}
   \begin{center}
      \includegraphics[width=0.95\linewidth]{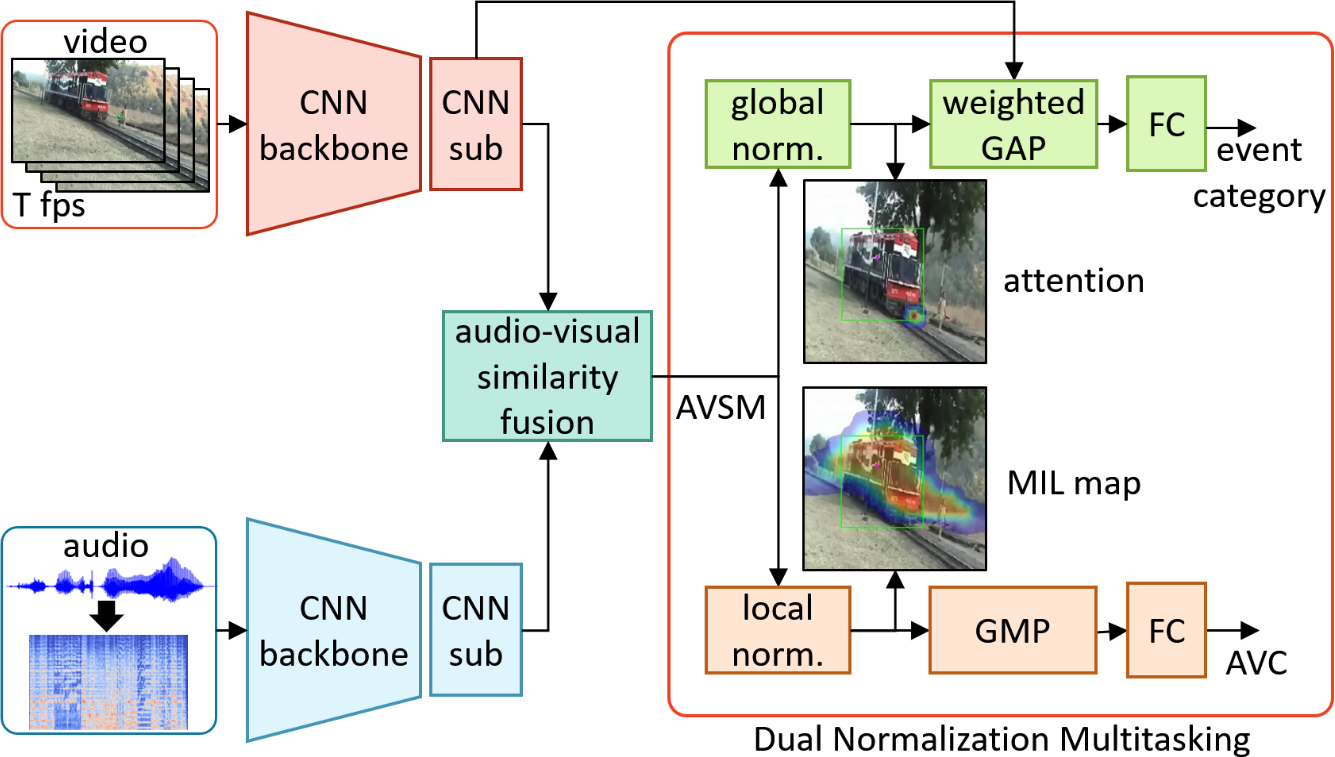}
   \end{center}
   \caption{
      Proposed architecture with Dual Normalization Multitasking (DNM).
      Video is converted to sequential images and two CNNs (backbone and sub) extract visual features.
      Video is converted to sequential images and two CNNs (backbone and sub) extract visual features.
      Audio waveform is converted to log-mel spectrogram and sound features are extracted by two CNNs.
      These features are fused based on similarity in pixel-wise mannar and the AVSM is obtained.
      The AVSM is normalized in two ways, globally and locally.
      Globally normalized map is used as an attention for the visual feature to solve event classification.
      Each pixel of locally normalized map is dealt as an instance of MIL to solve AVC.
      This simple strategy allows to directly connect the two training tasks into a single map for the purpose of localization, and effectively boost the performance.
   }
   \label{fig:archDNM}
\end{figure}

\begin{figure*}[t!]
   \begin{center}
      \begin{minipage}{0.49\hsize}
         \begin{center}
            \includegraphics[height=3.4cm]{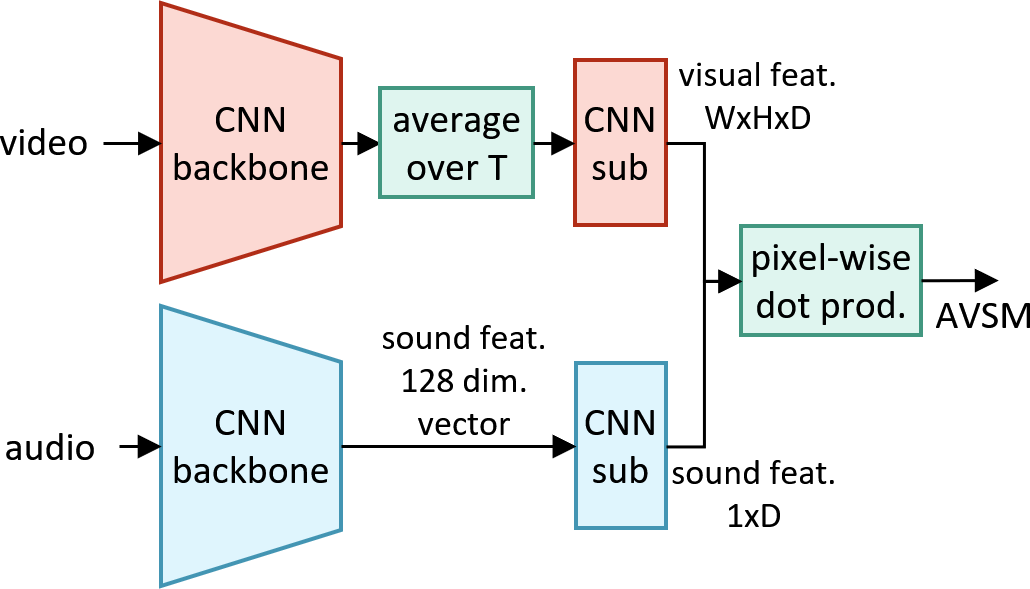}\\
            (a) Static Fusion
         \end{center}
      \end{minipage}
      \begin{minipage}{0.49\hsize}
         \begin{center}
            \includegraphics[height=3.4cm]{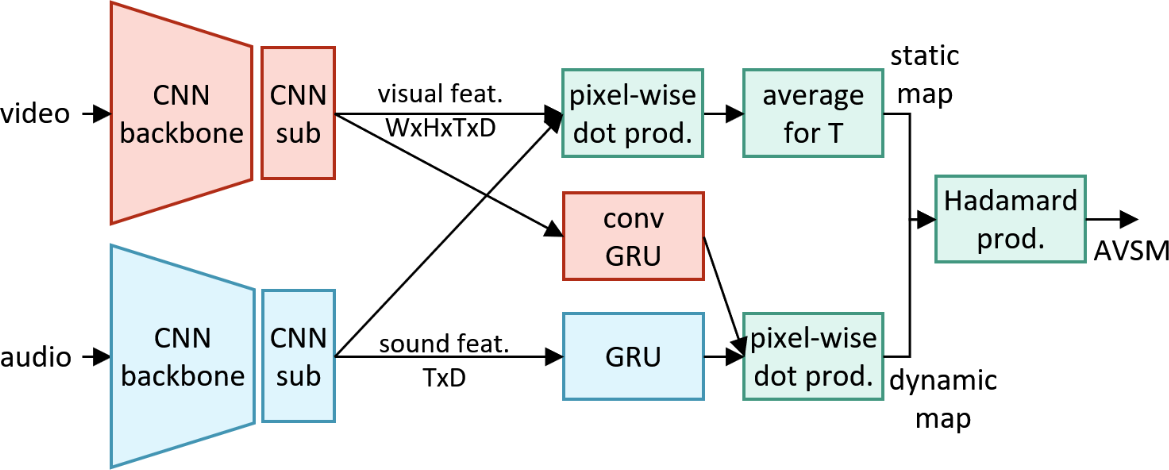}\\
            (b) Combined Dynamic Fusion (CDF)
         \end{center}
      \end{minipage}
   \end{center}
   \caption{
      Audio-Visual Similarity Fusion.
      (a) In the static fusion, we follow \cite{tian2018audio} for feature extraction.
      After the visual backbone, the feature is averaged to crush the dimension of time $T$.
      In audio, the last layer output of the VGGish\cite{hershey2017cnn} is used.
      AVSM is obtained by pixel-wise dot product without taking care of $T$.
      (b) The CDF is to capture the relation of detailed temporal changes between audio and video.
      The visual and sound features are fused in two ways, statically and dynamically with GRU\cite{cho2014learning}\cite{ballas2016delving}, and further combined to get the AVSM.
      The map is expected to represent the pixel-wise locations where video and audio have static correspondences (\ie visual appearances to sound types) and dynamic relations (\ie visual motions to sound variations) at the same time.
   }
   \label{fig:archSDSF}
\end{figure*}

\section{Proposed Algorithm}
As shown in Figure \ref{fig:archDNM}, our proposed AVSOL algorithm consists of the following sequences of steps:
(1) Extraction of video and audio features by CNN;
(2) Create pixel-wise AVSM by the fusion of both features;
(3) Normalize the AVSM in two different ways and connect to AVC task and event classification task respectively (DNM).

\subsection{Notations}
Following \cite{tian2018audio}, we split a video into non-overlapping segments of one second length each.
The event category in AVE Dataset is annotated at second-level.
The video and audio input of each one second segment are denoted as $V$ and $A$, respectively.
Note that a single video segment may contain events of multiple categories at the same time, but the label is only attached to one event in AVE dataset.

\subsection{Visual Feature Network}
The visual feature network consists of a pretrained backbone network and a sub network.
It has been shown that pretrained CNN features (VGG19 or I3D in our experiments) by a large-scale dataset are generic for other tasks.
A video segment $V$ is converted into a sequential RGB images, and the backbone extracts the visual features of them.

In order to preserve the spatial dimension, we use up to the middle layer of backbone where the spatial size of the features is $W \times H$.
The extracted features are further input to a sub network using 3D CNN to obtain the 4D visual feature
$v \in {\mathbb R}^{I \times T \times D}$,
where $I (=W \times H)$ is the vectorized spatial dimension of each feature map, $T$ denotes the time dimension and $D$ denotes the dimension of the feature vector.
Note that, unlike \cite{tian2018audio} or \cite{ramaswamy2020see}, the visual features extracted from one second segment retain the temporal dimensionality $T$.

\subsection{Audio Feature Network}
The audio feature network is used to extract audio features from one second raw monaural audio $A$.
Like the visual feature network, the audio feature network consists of a backbone network and a sub network.

The log mel spectrogram of audio, which can be treated as a single channel image, is input to the pretrained backbone CNN to extract features.
In order to preserve the temporal information, features with the spatial size of $B \times T$ are extracted using up to the middle layer of the CNN.
$B$ and  $T$  denotes the dimension of Mel Bin and time respectively.
They are further input to the sub network of 2D CNN to extract the audio features
$a \in {\mathbb R}^{T \times D}$,
where $D$ represents the dimension of the audio feature vector.

\subsection{Audio-Visual Similarity Fusion}
The features extracted from each audio and visual feature network are fused to create pixel-wise AVSM.
We propose two fusion methods: the static fusion and the Combined Dynamic Fusion (CDF).
Various types of audio and video fusion have been proposed so far\cite{tian2018audio,ephrat2018looking,ramaswamy2020see,owens2018audio}.
However in sound localization, previous works \cite{tian2018audio,ramaswamy2020see} did not look at fine temporal variations within one second.
These fusion methods may also be considered a type of static fusion.

\noindent
{\bf Static Fusion}
The static fusion, which serves as a baseline, simply takes the similarity between static appearance and sound feature.
As described in Figure \ref{fig:archSDSF} (a), we use the 1D vector audio feature of final layer output of audio backbone.

\noindent
{\bf Combined Dynamic Fusion}
From the findings in speaker estimation\cite{ephrat2018looking} and sound separation\cite{zhao2019sound}, we believe it is important to extract the dynamic correlation between audio and video for AVSOL.
For example, the movement of the hand playing an instrument and the change in the music sound will be highly correlated.
At the same time, as the mechanism of sound generation varies\cite{gaver1993world}, even objects with no apparent movement, such as car engines, produce sound.
Therefore, we propose CDF (Figure \ref{fig:archSDSF} (b)) which uses both static and dynamic features to compute the similarity between audio and video features.

Let $v_t\in {\mathbb R}^{I \times D}$ be the sliced visual feature $v$ along the time $T$, and $v_t^i \in {\mathbb R}^D$ be the feature vector at each pixel on $v_t$, \ie $v=\{v_t\}_{t=1}^T$ and $v_t=\{v_t^i\}_{i=1}^I$.
For audio features, we have $a=\{a_t\}_{t=1}^T$ in the same manner.
$a_t \in {\mathbb R}^D$ denotes the audio feature vector of the time period $t$.

For a one second clip, a static map is obtained as the time average of the pixel-wise dot product as
$M_{\rm Static} = \{{\rm average}_t(v_t^i \cdot a_t) | i=1,...,I\} \in {\mathbb R}^I.$
To make a dynamic map, each feature $v$ and $a$ is encoded by GRU\cite{cho2014learning} and then pixel-wise dot product is computed for the final output of each GRU.
For $v$, we use convolutional GRU (ConvGRU)\cite{ballas2016delving} since it has a spatial dimension.
ConvGRU and GRU takes $v_t$ and $a_t$ as input to encode temporal dependencies by processing them in unidirectional manner, for the two modalities respectively:
\begin{eqnarray}
   v'_T, h^v_T  & = & {\rm ConvGRU}(v_0, h^v_{0}) \\
   a'_T, h^a_T & = & {\rm GRU} (a_0, h^a_{0})
\end{eqnarray}
where  $v'_T \in {\mathbb R}^{I \times D}$ and $a'_T \in {\mathbb R}^{D}$ refer to final output of visual and sound feature and $h^v$ and $h^a$ represent hidden states.
The dynamic similarity is computed from  $v'_T$ and $a'_T$ as
$M_{\rm Dynamic} = \{{v'}_T^{i} \cdot a'_T) | i=1,...,I\} \in {\mathbb R}^I$.
By taking Hadamard product of static and dynamic maps, we obtain the final AVSM $M_{\rm CDF} \in {\mathbb R}^I$ as follows.
\begin{equation}
   M_{\rm CDF} = {M_{\rm Static}} \circ  {M_{\rm Dynamic}}
\end{equation}

\subsection{Dual Normalization Multitasking}
Our model is trained using both the AVC task and the event classification task.
The advantage of the AVC is that it is completely self-supervised.
It is also compatible with MIL formulation, as it is positive/negative classification.
The disadvantage is that the audio is often mixed with out-of-view sounds or ambient sounds that is not directly related to the visible events.
In fact, based on our experience with manually annotating AVSOL-E dataset, it is sometimes difficult even for expert human annotators to determine whether a certain sound comes from the object in the screen or not.

We believe this drawback can be compensated for by additional supervision, \eg the category of events shown in the video.
Since category annotations are made by humans, it's guaranteed the audio contains the sound from the target object.
The disadvantage is a request for manual annotation, but category annotations are much easier to correct than other types of annotations\cite{bearman2016s}.

Each of AVC and classification is commonly used in sound localization, but we don't yet have a known strategy on how to utilize them together to improve localization.
Therefore, we propose a method to effectively connect them: Dual Normalization Multitasking (DNM).

In DNM, we normalize the AVSM in two ways; locally and globally.
In local normalization, each pixel of the similarity map is processed independently.
Each normalized value is used as an instance for MIL to estimate AVC.
Specifically, sigmoid function is used for normalization, and then global max pooling (GMP) is applied to the entire map to produce the AVC estimation:
\begin{equation}
   z_{\rm avc} = \rm{GMP}(M_{\rm loc})
\end{equation}
where
$M_{\rm loc} = \{ \mathit{Sigmoid}(s_i)\}_{i=1}^I$
is the localization map and $s_i$ is the similarity score for each pixel in AVSM.
$M_{\rm loc}$ is used as the localization result of AVSOL.

The purpose of global normalization is to select important pixels by considering their relation to the entire AVSM.
We use the globally normalized map as the attention to visual features.
Specifically, softmax function is used to generate attention weights, which will be applied to obtain a weighted sum of the feature vectors $v_{\rm att} \in {\mathbb R}^{D'}$:
\begin{equation}
   v_{\rm att} = \Sigma_{i=1}^I w_{\rm att}^i v_{\rm cls}^i
\end{equation}
where $v_{\rm cls} \in {\mathbb R}^{I \times D'}$ is an output from the branch of sub CNN, and
$w_{\rm att}  =  \mathit{Softmax}(\{s_i\}_{i=1}^I) \in {\mathbb R}^{I}$ is attention obtained by global normalization.
Finally, $v_{\rm att}$ is passed to fully connected layer to obtain classification output.

\section{Experiments}
\subsection{Implementation Details}
For the AVC, we make positive (AVC=yes) and negative (AVC=no) data.
The positive data is the correct combination of audio and video.
For the negative data, the video and audio are separated, and the video is combined with the audio of a different clip or time.
The number of positive and negative data is set to 1:1.
For each training epoch, the combination of video and audio in the negative data is shuffled.

During training, the error of event classification is back-propagated only when AVC=yes.
Random horizontal flip and resized crop are used for video data augmentation.
No data augmentation is applied for audio.
We used a binary cross-entropy loss for both AVC and event classification, and trained our network with the sum of these two losses.

For the visual backbone, we compare VGG19\cite{SimonyanZ14a} pretrained with ImageNet\cite{imagenet_cvpr09}, and I3D's RGB stream \cite{carreira2017quo} pretrained with Kinetics Dataset\cite{carreira2017quo}.
VGG19 is commonly used in image recognition, and I3D is commonly used in video action recognition.
VGGish\cite{hershey2017cnn} pretrained by Audioset\cite{gemmeke2017audio} is used to extract audio features.
During training, the parameters of the backbone networks are fixed.

\subsection{Baselines and Evaluation Method}
We set two baselines for model comparison; CAM \cite{zhou2016learning} and the model proposed by Tian \etal\cite{tian2018audio}.
CAM is a basic and powerful method in WSOL.
We modified our proposed model to take only visual input and output CAM.
Specifically, we applied global average pooling (GAP) to the feature $v_{\rm cls}$ from visual sub CNN followed by the fully connected layer to make CAM.

In \cite{tian2018audio}, the audio-guided visual attention is used to localize sounding objects.
We used the model trained and provided by the authors.
With referring to WSOL evaluation, we normalized the output attention by min-max for each frame individually\cite{choe2020evaluating}.
\cite{tian2018audio} uses whole 10-second video as input, and their temporal interrelationships for 10 seconds are coded in bidirectional LSTM.
However, we humans can recognize the type of event and the location of sound source instantly and in real-time\cite{gaver1993world}.
Since our algorithm is intended to be used in real-world and real-time applications, we use only 1-second videos as input to solve the problems.
Note that our algorighm solves tasks with much less information than Tian \etal's\cite{tian2018audio} and other methods that follow\cite{ramaswamy2020see, duan2021audio, wu2019dual, lin2019dual}.

In addition to the two baselines, to validate the effectiveness of DNM, we compared models trained on a single task, either AVC or event classification, to the model trained using DNM.
For AVC and DNM, the MIL map $M_{\rm loc}$ is used as the localization result.
For the classification only models, the attention weight $w_{\rm att}$ is used as localization result.

HmBoxAUC, PiBR, and PNSR are used for evaluation.
We choose the model with the maximum accuracy in AVC and classification in the validation set respectively, and evaluate them for AVSOL in the test set.
Multiple trainings are conducted and the average result is reported.

\subsection{experimental comparisons}

\begin{table*}
   \begin{center}
      \begin{tabular}{lcccccccccc}
         \toprule
         \multirow{2}{*}{Models} & \multicolumn{3}{c}{HmBoxAUC$\uparrow$} & \multicolumn{3}{c}{PiBR$\uparrow$} & \multicolumn{4}{c}{PNSR$\downarrow$}                                                                                                   \\ \cmidrule(rl){2-4} \cmidrule(rl){5-7} \cmidrule(rl){8-11}
                                 & {\bf all}                              & single                             & multi.                               & {\bf all}   & single      & multi.      & {\bf all}   & visible     & audible     & noise       \\
         \midrule
         CAM\cite{zhou2016learning} (visual only)
                                 & 0.416                                  & 0.433                              & 0.348                                & 0.629       & 0.625       & 0.644       & -           & -           & -           & -           \\
         Tian \etal\cite{tian2018audio} 
                                 & 0.508                                  & 0.531                              & 0.432                                & 0.671       & 0.687       & 0.617       & -           & -           & -           & -           \\
         \midrule
         VGG19-Static-Cls.       & 0.049                                  & 0.046                              & 0.061                                & 0.672       & 0.691       & 0.604       & -           & -           & -           & -           \\
         VGG19-Static-AVC        & 0.328                                  & 0.350                              & 0.250                                & 0.450       & 0.484       & 0.331       & 0.805       & 0.881       & 0.816       & 0.317       \\
         VGG19-Static-DNM        & 0.571                                  & 0.604                              & 0.419                                & 0.630       & 0.667       & 0.550       & 0.660       & 0.610       & 0.701       & 0.298       \\
         VGG19-CDF-DNM           & 0.573                                  & 0.598                              & {\bf 0.467}                          & 0.672       & 0.688       & 0.617       & 0.605       & 0.677       & 0.619       & 0.307       \\
         \midrule
         I3D-Static-Cls.         & 0.278                                  & 0.279                              & 0.292                                & 0.661       & 0.672       & 0.620       & -           & -           & -           & -           \\
         I3D-Static-AVC          & 0.480                                  & 0.499                              & 0.414                                & 0.717       & 0.746       & 0.615       & 0.590       & 0.679       & 0.604       & 0.268       \\
         I3D-Static-DNM          & 0.576                                  & {\bf 0.619}                        & 0.422                                & 0.709       & 0.739       & 0.602       & 0.626       & 0.637       & 0.658       & {\bf 0.241} \\
         I3D-CDF-DNM             & {\bf 0.581}                            & 0.616                              & 0.451                                & {\bf 0.744} & {\bf 0.768} & {\bf 0.659} & {\bf 0.550} & {\bf 0.587} & {\bf 0.563} & 0.328       \\
         \bottomrule
      \end{tabular}
   \end{center}
   \caption{
      Comparisons of our models and baselines on AVSOL-E dataset.
      HmBoxAUC and PiBR are evaluated for all the AVE frames (all).
      They are also separately evaluated for the scenes with single sounding object (single) and for the scenes with multiple sounding objects (multi.).
      PNSR for all the frames and three separated conditions (visible, audible and noise) are reported for the models with AVC output.
   }
   \label{table:result}
\end{table*}

\begin{figure*}
   \begin{center}
      \begin{minipage}{0.99\linewidth}
         \begin{minipage}{0.091\linewidth}
            \begin{center}
               {\scriptsize Annotations}
            \end{center}
         \end{minipage}
         \hspace{-4pt}
         \begin{minipage}{0.091\linewidth}
            \begin{center}
               {\scriptsize CAM}
            \end{center}
         \end{minipage}
         \hspace{-4pt}
         \begin{minipage}{0.091\linewidth}
            \begin{center}
               {\scriptsize Tian \etal}
            \end{center}
         \end{minipage}
         \hspace{-4pt}
         \begin{minipage}{0.091\linewidth}
            \begin{center}
               {\scriptsize VGG19}\\
               \vspace{-1ex}
               {\scriptsize Static-Cls.}
            \end{center}
         \end{minipage}
         \hspace{-4pt}
         \begin{minipage}{0.091\linewidth}
            \begin{center}
               {\scriptsize VGG19}\\
               \vspace{-1ex}
               {\scriptsize Static-AVC}
            \end{center}
         \end{minipage}
         \hspace{-4pt}
         \begin{minipage}{0.091\linewidth}
            \begin{center}
               {\scriptsize VGG19}\\
               \vspace{-1ex}
               {\scriptsize Static-DNM}
            \end{center}
         \end{minipage}
         \hspace{-4pt}
         \begin{minipage}{0.091\linewidth}
            \begin{center}
               {\scriptsize VGG19}\\
               \vspace{-1ex}
               {\scriptsize CDF-DNM}
            \end{center}
         \end{minipage}
         \hspace{-4pt}
         \begin{minipage}{0.091\linewidth}
            \begin{center}
               {\scriptsize I3D}\\
               \vspace{-1ex}
               {\scriptsize Static-Cls.}
            \end{center}
         \end{minipage}
         \hspace{-4pt}
         \begin{minipage}{0.091\linewidth}
            \begin{center}
               {\scriptsize I3D}\\
               \vspace{-1ex}
               {\scriptsize Static-AVC}
            \end{center}
         \end{minipage}
         \hspace{-4pt}
         \begin{minipage}{0.091\linewidth}
            \begin{center}
               {\scriptsize I3D}\\
               \vspace{-1ex}
               {\scriptsize Static-DNM}
            \end{center}
         \end{minipage}
         \hspace{-4pt}
         \begin{minipage}{0.091\linewidth}
            \begin{center}
               {\scriptsize I3D}\\
               \vspace{-1ex}
               {\scriptsize CDF-DNM}
            \end{center}
         \end{minipage}
      \end{minipage}

      \begin{minipage}{0.99\linewidth}
         \begin{minipage}{0.091\linewidth}
            \begin{center}
               \includegraphics[width=\linewidth]{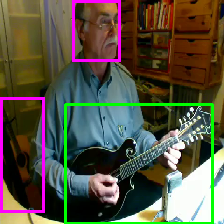}
            \end{center}
         \end{minipage}
         \hspace{-4pt}
         \begin{minipage}{0.091\linewidth}
            \begin{center}
               \includegraphics[width=\linewidth]{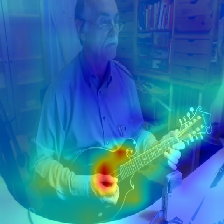}
            \end{center}
         \end{minipage}
         \hspace{-4pt}
         \begin{minipage}{0.091\linewidth}
            \begin{center}
               \includegraphics[width=\linewidth]{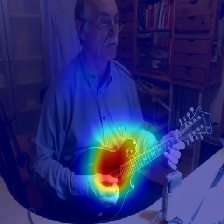}
            \end{center}
         \end{minipage}
         \hspace{-4pt}
         \begin{minipage}{0.091\linewidth}
            \begin{center}
               \includegraphics[width=\linewidth]{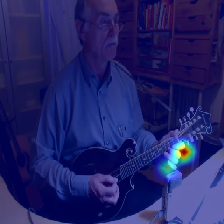}
            \end{center}
         \end{minipage}
         \hspace{-4pt}
         \begin{minipage}{0.091\linewidth}
            \begin{center}
               \includegraphics[width=\linewidth]{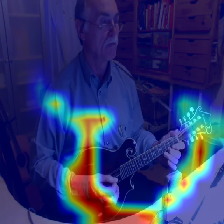}
            \end{center}
         \end{minipage}
         \hspace{-4pt}
         \begin{minipage}{0.091\linewidth}
            \begin{center}
               \includegraphics[width=\linewidth]{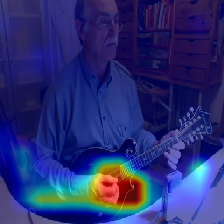}
            \end{center}
         \end{minipage}
         \hspace{-4pt}
         \begin{minipage}{0.091\linewidth}
            \begin{center}
               \includegraphics[width=\linewidth]{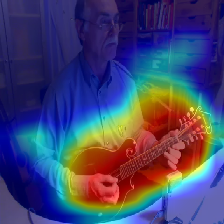}
            \end{center}
         \end{minipage}
         \hspace{-4pt}
         \begin{minipage}{0.091\linewidth}
            \begin{center}
               \includegraphics[width=\linewidth]{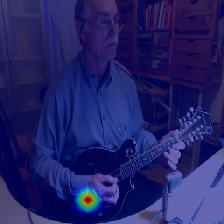}
            \end{center}
         \end{minipage}
         \hspace{-4pt}
         \begin{minipage}{0.091\linewidth}
            \begin{center}
               \includegraphics[width=\linewidth]{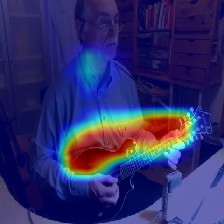}
            \end{center}
         \end{minipage}
         \hspace{-4pt}
         \begin{minipage}{0.091\linewidth}
            \begin{center}
               \includegraphics[width=\linewidth]{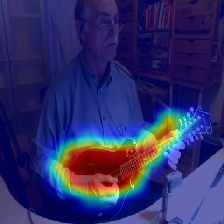}
            \end{center}
         \end{minipage}
         \hspace{-4pt}
         \begin{minipage}{0.091\linewidth}
            \begin{center}
               \includegraphics[width=\linewidth]{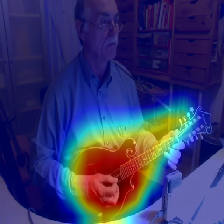}
            \end{center}
         \end{minipage}
      \end{minipage}

      \begin{minipage}{0.99\linewidth}
         \begin{minipage}{0.091\linewidth}
            \begin{center}
               \includegraphics[width=\linewidth]{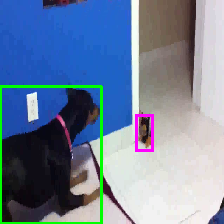}
            \end{center}
         \end{minipage}
         \hspace{-4pt}
         \begin{minipage}{0.091\linewidth}
            \begin{center}
               \includegraphics[width=\linewidth]{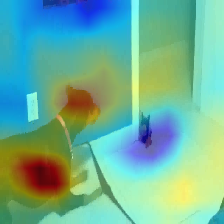}
            \end{center}
         \end{minipage}
         \hspace{-4pt}
         \begin{minipage}{0.091\linewidth}
            \begin{center}
               \includegraphics[width=\linewidth]{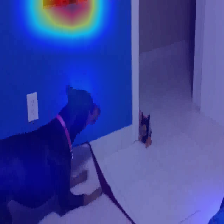}
            \end{center}
         \end{minipage}
         \hspace{-4pt}
         \begin{minipage}{0.091\linewidth}
            \begin{center}
               \includegraphics[width=\linewidth]{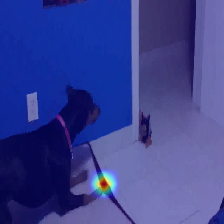}
            \end{center}
         \end{minipage}
         \hspace{-4pt}
         \begin{minipage}{0.091\linewidth}
            \begin{center}
               \includegraphics[width=\linewidth]{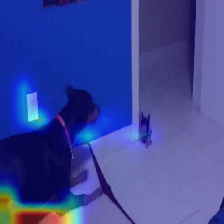}
            \end{center}
         \end{minipage}
         \hspace{-4pt}
         \begin{minipage}{0.091\linewidth}
            \begin{center}
               \includegraphics[width=\linewidth]{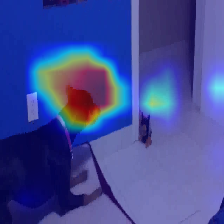}
            \end{center}
         \end{minipage}
         \hspace{-4pt}
         \begin{minipage}{0.091\linewidth}
            \begin{center}
               \includegraphics[width=\linewidth]{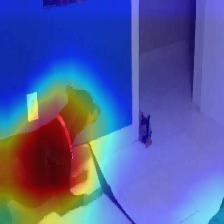}
            \end{center}
         \end{minipage}
         \hspace{-4pt}
         \begin{minipage}{0.091\linewidth}
            \begin{center}
               \includegraphics[width=\linewidth]{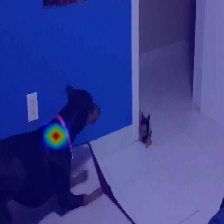}
            \end{center}
         \end{minipage}
         \hspace{-4pt}
         \begin{minipage}{0.091\linewidth}
            \begin{center}
               \includegraphics[width=\linewidth]{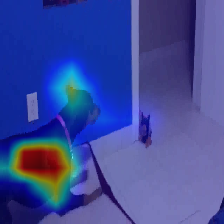}
            \end{center}
         \end{minipage}
         \hspace{-4pt}
         \begin{minipage}{0.091\linewidth}
            \begin{center}
               \includegraphics[width=\linewidth]{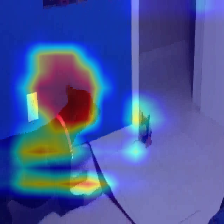}
            \end{center}
         \end{minipage}
         \hspace{-4pt}
         \begin{minipage}{0.091\linewidth}
            \begin{center}
               \includegraphics[width=\linewidth]{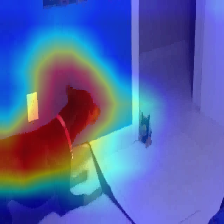}
            \end{center}
         \end{minipage}
      \end{minipage}

      \begin{minipage}{0.99\linewidth}
         \begin{minipage}{0.091\linewidth}
            \begin{center}
               \includegraphics[width=\linewidth]{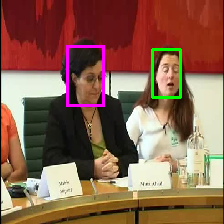}
            \end{center}
         \end{minipage}
         \hspace{-4pt}
         \begin{minipage}{0.091\linewidth}
            \begin{center}
               \includegraphics[width=\linewidth]{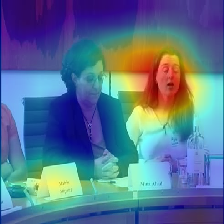}
            \end{center}
         \end{minipage}
         \hspace{-4pt}
         \begin{minipage}{0.091\linewidth}
            \begin{center}
               \includegraphics[width=\linewidth]{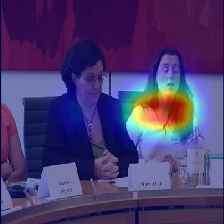}
            \end{center}
         \end{minipage}
         \hspace{-4pt}
         \begin{minipage}{0.091\linewidth}
            \begin{center}
               \includegraphics[width=\linewidth]{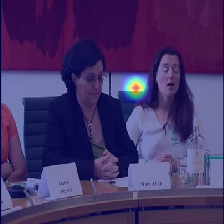}
            \end{center}
         \end{minipage}
         \hspace{-4pt}
         \begin{minipage}{0.091\linewidth}
            \begin{center}
               \includegraphics[width=\linewidth]{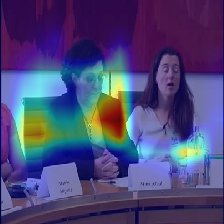}
            \end{center}
         \end{minipage}
         \hspace{-4pt}
         \begin{minipage}{0.091\linewidth}
            \begin{center}
               \includegraphics[width=\linewidth]{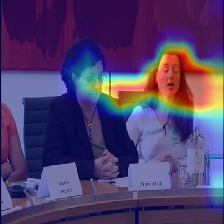}
            \end{center}
         \end{minipage}
         \hspace{-4pt}
         \begin{minipage}{0.091\linewidth}
            \begin{center}
               \includegraphics[width=\linewidth]{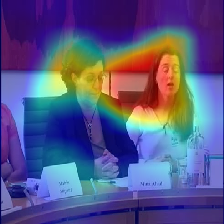}
            \end{center}
         \end{minipage}
         \hspace{-4pt}
         \begin{minipage}{0.091\linewidth}
            \begin{center}
               \includegraphics[width=\linewidth]{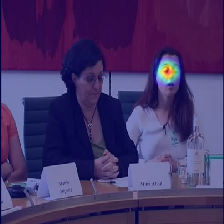}
            \end{center}
         \end{minipage}
         \hspace{-4pt}
         \begin{minipage}{0.091\linewidth}
            \begin{center}
               \includegraphics[width=\linewidth]{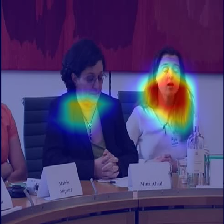}
            \end{center}
         \end{minipage}
         \hspace{-4pt}
         \begin{minipage}{0.091\linewidth}
            \begin{center}
               \includegraphics[width=\linewidth]{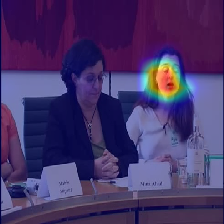}
            \end{center}
         \end{minipage}
         \hspace{-4pt}
         \begin{minipage}{0.091\linewidth}
            \begin{center}
               \includegraphics[width=\linewidth]{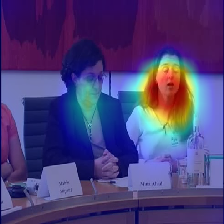}
            \end{center}
         \end{minipage}
      \end{minipage}

      \begin{minipage}{0.99\linewidth}
         \begin{minipage}{0.091\linewidth}
            \begin{center}
               \includegraphics[width=\linewidth]{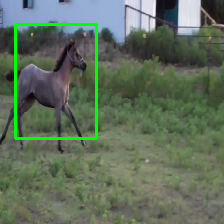}
            \end{center}
         \end{minipage}
         \hspace{-4pt}
         \begin{minipage}{0.091\linewidth}
            \begin{center}
               \includegraphics[width=\linewidth]{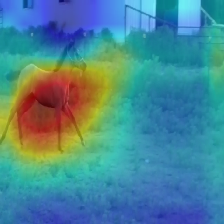}
            \end{center}
         \end{minipage}
         \hspace{-4pt}
         \begin{minipage}{0.091\linewidth}
            \begin{center}
               \includegraphics[width=\linewidth]{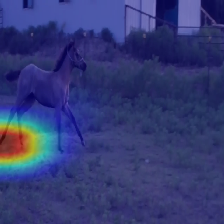}
            \end{center}
         \end{minipage}
         \hspace{-4pt}
         \begin{minipage}{0.091\linewidth}
            \begin{center}
               \includegraphics[width=\linewidth]{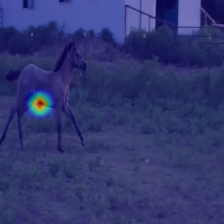}
            \end{center}
         \end{minipage}
         \hspace{-4pt}
         \begin{minipage}{0.091\linewidth}
            \begin{center}
               \includegraphics[width=\linewidth]{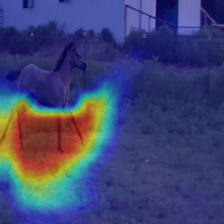}
            \end{center}
         \end{minipage}
         \hspace{-4pt}
         \begin{minipage}{0.091\linewidth}
            \begin{center}
               \includegraphics[width=\linewidth]{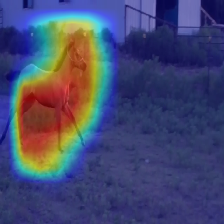}
            \end{center}
         \end{minipage}
         \hspace{-4pt}
         \begin{minipage}{0.091\linewidth}
            \begin{center}
               \includegraphics[width=\linewidth]{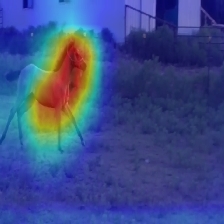}
            \end{center}
         \end{minipage}
         \hspace{-4pt}
         \begin{minipage}{0.091\linewidth}
            \begin{center}
               \includegraphics[width=\linewidth]{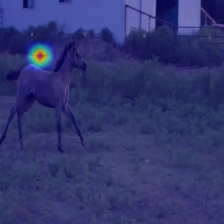}
            \end{center}
         \end{minipage}
         \hspace{-4pt}
         \begin{minipage}{0.091\linewidth}
            \begin{center}
               \includegraphics[width=\linewidth]{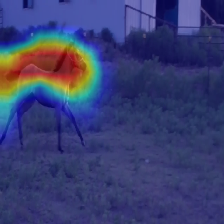}
            \end{center}
         \end{minipage}
         \hspace{-4pt}
         \begin{minipage}{0.091\linewidth}
            \begin{center}
               \includegraphics[width=\linewidth]{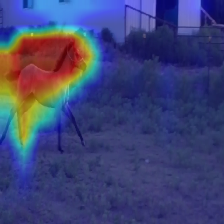}
            \end{center}
         \end{minipage}
         \hspace{-4pt}
         \begin{minipage}{0.091\linewidth}
            \begin{center}
               \includegraphics[width=\linewidth]{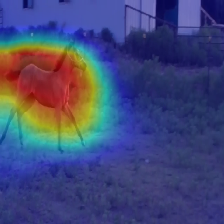}
            \end{center}
         \end{minipage}
      \end{minipage}

      \begin{minipage}{0.99\linewidth}
         \begin{minipage}{0.091\linewidth}
            \begin{center}
               \includegraphics[width=\linewidth]{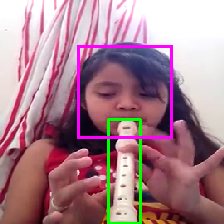}
            \end{center}
         \end{minipage}
         \hspace{-4pt}
         \begin{minipage}{0.091\linewidth}
            \begin{center}
               \includegraphics[width=\linewidth]{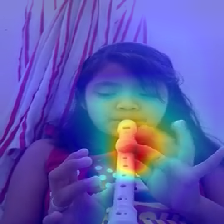}
            \end{center}
         \end{minipage}
         \hspace{-4pt}
         \begin{minipage}{0.091\linewidth}
            \begin{center}
               \includegraphics[width=\linewidth]{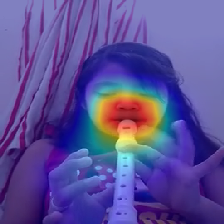}
            \end{center}
         \end{minipage}
         \hspace{-4pt}
         \begin{minipage}{0.091\linewidth}
            \begin{center}
               \includegraphics[width=\linewidth]{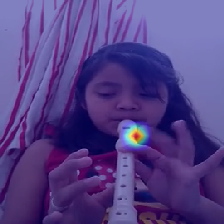}
            \end{center}
         \end{minipage}
         \hspace{-4pt}
         \begin{minipage}{0.091\linewidth}
            \begin{center}
               \includegraphics[width=\linewidth]{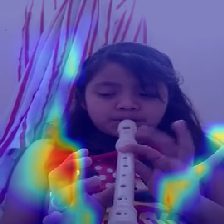}
            \end{center}
         \end{minipage}
         \hspace{-4pt}
         \begin{minipage}{0.091\linewidth}
            \begin{center}
               \includegraphics[width=\linewidth]{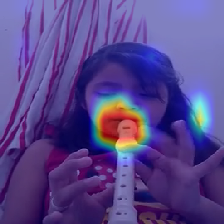}
            \end{center}
         \end{minipage}
         \hspace{-4pt}
         \begin{minipage}{0.091\linewidth}
            \begin{center}
               \includegraphics[width=\linewidth]{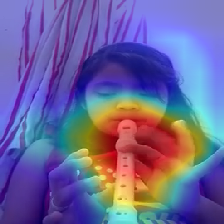}
            \end{center}
         \end{minipage}
         \hspace{-4pt}
         \begin{minipage}{0.091\linewidth}
            \begin{center}
               \includegraphics[width=\linewidth]{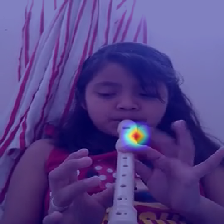}
            \end{center}
         \end{minipage}
         \hspace{-4pt}
         \begin{minipage}{0.091\linewidth}
            \begin{center}
               \includegraphics[width=\linewidth]{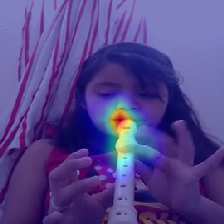}
            \end{center}
         \end{minipage}
         \hspace{-4pt}
         \begin{minipage}{0.091\linewidth}
            \begin{center}
               \includegraphics[width=\linewidth]{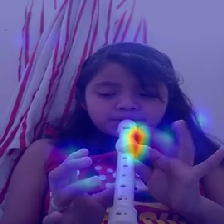}
            \end{center}
         \end{minipage}
         \hspace{-4pt}
         \begin{minipage}{0.091\linewidth}
            \begin{center}
               \includegraphics[width=\linewidth]{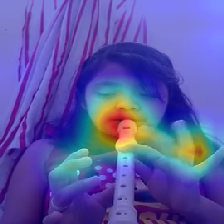}
            \end{center}
         \end{minipage}
      \end{minipage}

   \end{center}
   \caption{
      Qualitative comparisons of baselines and our models with AVSOL-E annotations.
      AVSOL-E annotations are shown in the leftmost column.
      Sounding objects (green bounding-box) and potential sounding objects which is not making sound at the frame (magenta bounding-box) are drawn.
      Heatmap is min-max normalized for better visibility.
   }
   \label{fig:qualitative}
\end{figure*}

\begin{figure}
   \begin{center}
      \begin{minipage}{\linewidth}

         \begin{minipage}{0.04\linewidth}
            \begin{center}
            \end{center}
         \end{minipage}
         \begin{minipage}{0.19\linewidth}
            \begin{center}
               {\scriptsize Annotations}
            \end{center}
         \end{minipage}
         \hspace{-5pt}
         \begin{minipage}{0.19\linewidth}
            \begin{center}
               {\scriptsize CAM}
            \end{center}
         \end{minipage}
         \hspace{-5pt}
         \begin{minipage}{0.19\linewidth}
            \begin{center}
               {\scriptsize Tian \etal}
            \end{center}
         \end{minipage}
         \hspace{-5pt}
         \begin{minipage}{0.19\linewidth}
            \begin{center}
               {\scriptsize VGG19}\\
               \vspace{-1ex}
               {\scriptsize CDF-DNM}
            \end{center}
         \end{minipage}
         \hspace{-5pt}
         \begin{minipage}{0.19\linewidth}
            \begin{center}
               {\scriptsize I3D}\\
               \vspace{-1ex}
               {\scriptsize CDF-DNM}
            \end{center}
         \end{minipage}

         \begin{minipage}{0.04\linewidth}
            \begin{center}
               {\footnotesize (a)}
            \end{center}
         \end{minipage}
         \begin{minipage}{0.19\linewidth}
            \begin{center}
               \includegraphics[width=\linewidth]{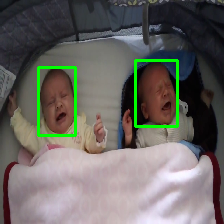}
            \end{center}
         \end{minipage}
         \hspace{-5pt}
         \begin{minipage}{0.19\linewidth}
            \begin{center}
               \includegraphics[width=\linewidth]{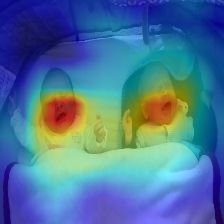}
            \end{center}
         \end{minipage}
         \hspace{-5pt}
         \begin{minipage}{0.19\linewidth}
            \begin{center}
               \includegraphics[width=\linewidth]{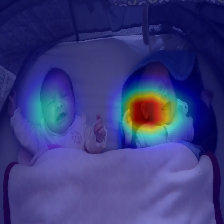}
            \end{center}
         \end{minipage}
         \hspace{-5pt}
         \begin{minipage}{0.19\linewidth}
            \begin{center}
               \includegraphics[width=\linewidth]{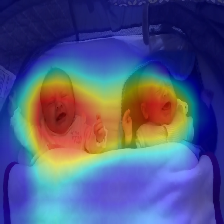}
            \end{center}
         \end{minipage}
         \hspace{-5pt}
         \begin{minipage}{0.19\linewidth}
            \begin{center}
               \includegraphics[width=\linewidth]{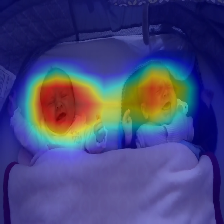}
            \end{center}
         \end{minipage}

         \begin{minipage}{0.04\linewidth}
            \begin{center}
               {\footnotesize (b)}
            \end{center}
         \end{minipage}
         \begin{minipage}{0.19\linewidth}
            \begin{center}
               \includegraphics[width=\linewidth]{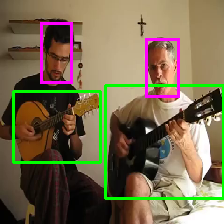}
            \end{center}
         \end{minipage}
         \hspace{-5pt}
         \begin{minipage}{0.19\linewidth}
            \begin{center}
               \includegraphics[width=\linewidth]{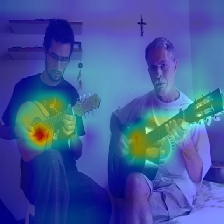}
            \end{center}
         \end{minipage}
         \hspace{-5pt}
         \begin{minipage}{0.19\linewidth}
            \begin{center}
               \includegraphics[width=\linewidth]{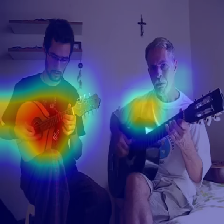}
            \end{center}
         \end{minipage}
         \hspace{-5pt}
         \begin{minipage}{0.19\linewidth}
            \begin{center}
               \includegraphics[width=\linewidth]{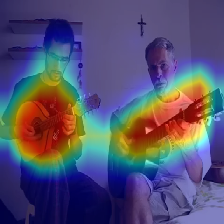}
            \end{center}
         \end{minipage}
         \hspace{-5pt}
         \begin{minipage}{0.19\linewidth}
            \begin{center}
               \includegraphics[width=\linewidth]{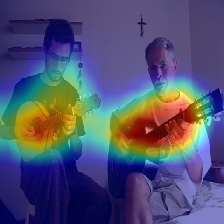}
            \end{center}
         \end{minipage}

         \begin{minipage}{0.04\linewidth}
            \begin{center}
               {\footnotesize (c)}
            \end{center}
         \end{minipage}
         \begin{minipage}{0.19\linewidth}
            \begin{center}
               \includegraphics[width=\linewidth]{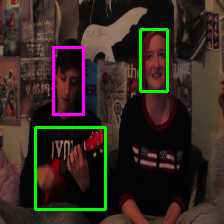}
            \end{center}
         \end{minipage}
         \hspace{-5pt}
         \begin{minipage}{0.19\linewidth}
            \begin{center}
               \includegraphics[width=\linewidth]{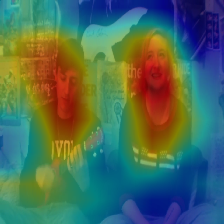}
            \end{center}
         \end{minipage}
         \hspace{-5pt}
         \begin{minipage}{0.19\linewidth}
            \begin{center}
               \includegraphics[width=\linewidth]{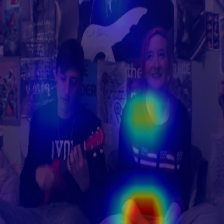}
            \end{center}
         \end{minipage}
         \hspace{-5pt}
         \begin{minipage}{0.19\linewidth}
            \begin{center}
               \includegraphics[width=\linewidth]{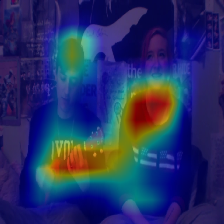}
            \end{center}
         \end{minipage}
         \hspace{-5pt}
         \begin{minipage}{0.19\linewidth}
            \begin{center}
               \includegraphics[width=\linewidth]{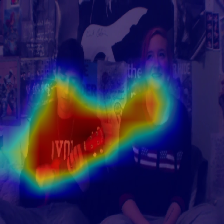}
            \end{center}
         \end{minipage}

         \begin{minipage}{0.04\linewidth}
            \begin{center}
               {\footnotesize (d)}
            \end{center}
         \end{minipage}
         \begin{minipage}{0.19\linewidth}
            \begin{center}
               \includegraphics[width=\linewidth]{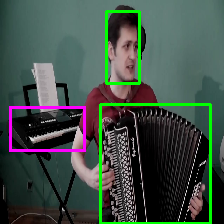}
            \end{center}
         \end{minipage}
         \hspace{-5pt}
         \begin{minipage}{0.19\linewidth}
            \begin{center}
               \includegraphics[width=\linewidth]{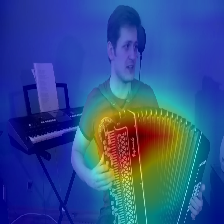}
            \end{center}
         \end{minipage}
         \hspace{-5pt}
         \begin{minipage}{0.19\linewidth}
            \begin{center}
               \includegraphics[width=\linewidth]{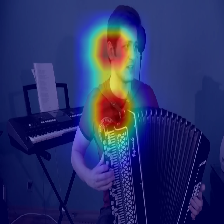}
            \end{center}
         \end{minipage}
         \hspace{-5pt}
         \begin{minipage}{0.19\linewidth}
            \begin{center}
               \includegraphics[width=\linewidth]{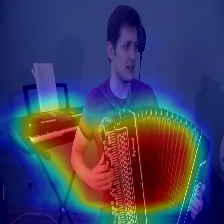}
            \end{center}
         \end{minipage}
         \hspace{-5pt}
         \begin{minipage}{0.19\linewidth}
            \begin{center}
               \includegraphics[width=\linewidth]{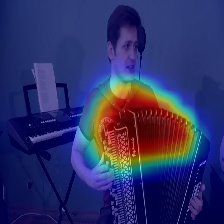}
            \end{center}
         \end{minipage}

      \end{minipage}
   \end{center}
   \caption{
      Comparisons in scenes with multiple sounding objects.
      Sounding objects (green bounding-box) and objects not making sound (magenta bounding-box) are visualized in annotations.
   }
   \label{fig:multiple}
\end{figure}

Table \ref{table:result} and Figure \ref{fig:qualitative} shows the quantitative and qualitative results respectively.
The result of two baseline methods and eight conditions of our proposed models are presented.

\noindent
{\bf Training Methods.}
For the models trained only with event classification (VGG19-Static-Cls, I3D-Static-Cls.), the HmBoxAUC are significantly lower than those of the baselines.
The qualitative results show that the globally normalized audio-visual attention map $wv_{\rm att}$ tends to concentrate on a point of the most discriminative region, and the shape of the target object is not extracted.
In spite of low HmBoxAUC, the PiBR is as good as those of baselines.

When trained with the AVC task only (VGG19-Static-AVC, I3D-Static-AVC), the performances are not as good as Tian \etal and
for VGG19-Static-AVC, the result is worse than I3D-Static-AVC and CAM.
The qualitative results show that the model tends to extract not the sounding object itself, but the area around its edges.
We saw this occur in most of the trials of training.
This appears to be same to the case reported by Senocak \etal\cite{senocak2018learning}.
For example, in music play, the visual feature of the edges of instrument which is almost always present on the screen at the same time with the instrument, may be learned to be more strongly associated with the sound.
This phenomenon is also known to be appear in the field of WSOL as the inherent ill-posedness of weakly-supervised learning \cite{choe2020evaluating} .

The two DNM models with static fusion (VGG19-Static-DNM, I3D-Static-DNM) showed a significant improvement in HmBoxAUC from the single-task conditions.
The qualitative results show that the models with DNM did not show the problems which occurred in single task conditions.
Since the MIL is good at extracting wide region, and the attention is good at finding globally important points, we think the combination of these two worked well.

\noindent
{\bf Audio-Visual Fusion Methods.}
The two CDF conditions (VGG19-CDF-DNM, I3D-CDF-DNM) showed better performance compared to their static fusion counterparts (VGG19-Static-DNM, I3D-Static-DNM) for all the three metrics.
The qualitative results show that the method is able to extract the entire area of the target object better than others (Figure \ref{fig:qualitative} top and 2nd row).
However, compared to static fusion, strong heatmap output sometimes appeared on the area of the moving parts.
As shown in Figure \ref{fig:qualitative} (bottom), moving fingers playing the flute are also localized.

\noindent
{\bf Multiple Sounding Objects.}
As shown in Table \ref{table:result}, in most cases, the HmBoxAUC and PiBR in single sounding object condition is better than that of multiple sounding objects.
Qualitative results in Figure \ref{fig:multiple} shows that both the baselines and the proposed methods with CDF and DNM succeed in localizing multiple sounding objects, when those objects are in the same or similar categories (Figure \ref{fig:multiple} (a) and (b)).
In such a case,
the model simply has to find the location where the visual feature corresponds to the sound feature.

It is more difficult to localize multiple sounding objects when they are in different categories.
The model has to represent two types of sound features together in a single sound feature vector, and associate them with different visual features in different positions.
Figure \ref{fig:multiple} (c) and (d) shows scenes with playing instrument and singing at the same time.
Our models only managed to locate the instrument being played, not the singing face.

\noindent
{\bf Non-AVE Scenes.}
\begin{figure}
   \begin{center}
      \begin{minipage}{0.98\linewidth}
         \begin{minipage}{0.05\linewidth}
            \begin{center}
               {\footnotesize (a)}
            \end{center}
         \end{minipage}
         \begin{minipage}{0.94\linewidth}
            \begin{center}
               \includegraphics[width=\linewidth]{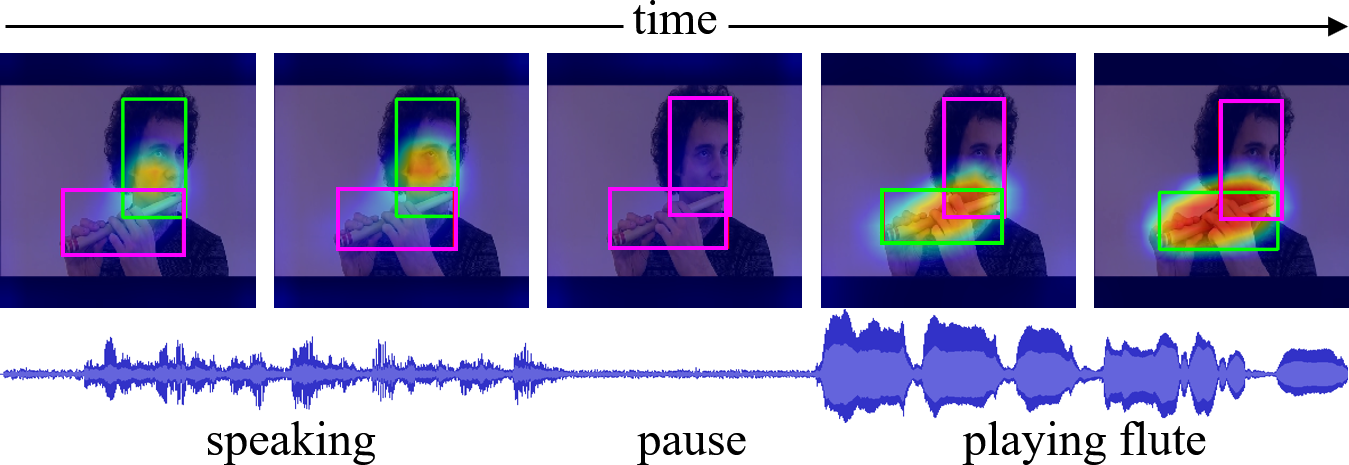}
            \end{center}
         \end{minipage}
      \end{minipage}

      \vspace{2mm}
      \begin{minipage}{0.98\linewidth}
         \begin{minipage}{0.05\linewidth}
            \begin{center}
               {\footnotesize (b)}
            \end{center}
         \end{minipage}
         \begin{minipage}{0.94\linewidth}
            \begin{center}
               \includegraphics[width=\linewidth]{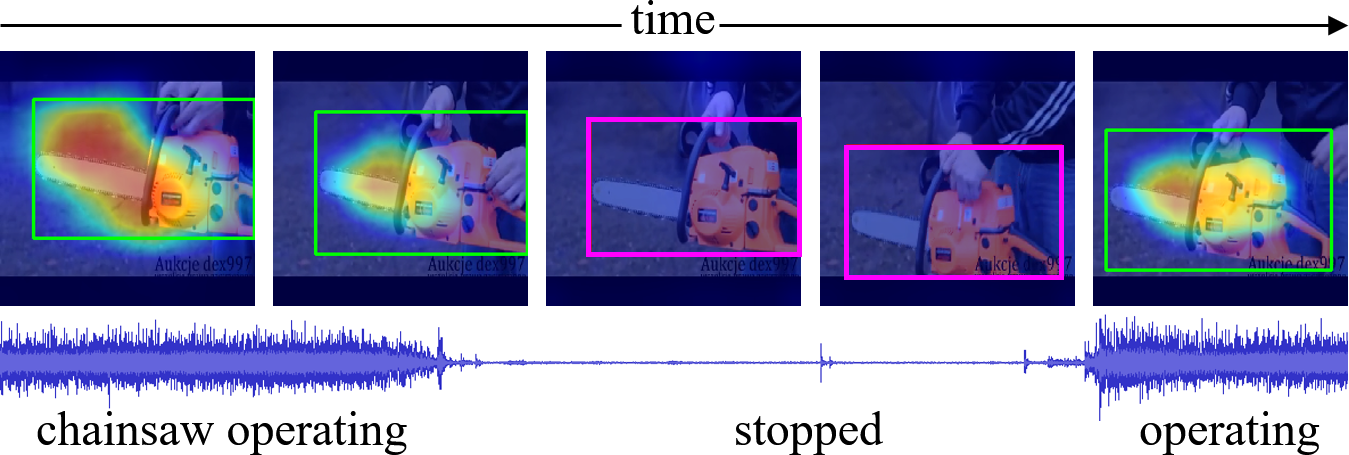}
            \end{center}
         \end{minipage}
      \end{minipage}

      \vspace{2mm}
      \begin{minipage}{0.98\linewidth}
         \begin{minipage}{0.05\linewidth}
            \begin{center}
               {\footnotesize (c)}
            \end{center}
         \end{minipage}
         \begin{minipage}{0.94\linewidth}
            \begin{center}
               \includegraphics[width=\linewidth]{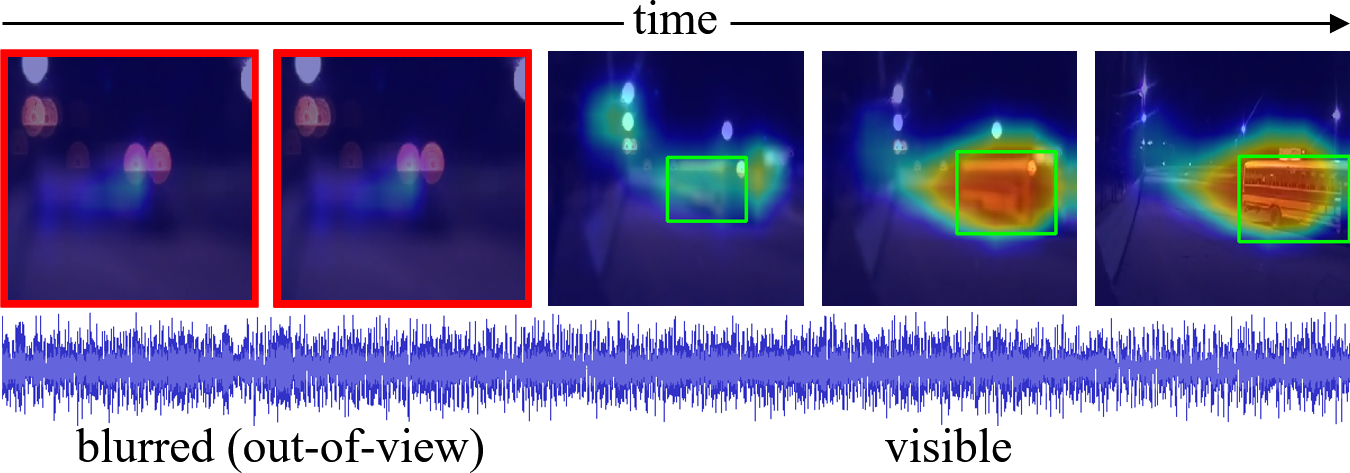}
            \end{center}
         \end{minipage}
      \end{minipage}
   \end{center}

   \caption{
      Qualitative results of a proposed model (I3D-CDF-DNM) for videos containing non-AVE scenes.
      Audio waveforms are also drawn.
      Frames that contain at least one sounding object (green bounding-box) are AVE frames.
      Frames only with objects not making sound (magenta bounding-box) or out-of-view sound (red bounding-box surrounding the frame) are non-AVE frames.
   }
   \label{fig:non-ave}
\end{figure}

PNSR for all different types of non-AVE scenes are shown in Table \ref{table:result}.
For all the tested models, in the scene with no sounding object nor no out-of-view sound (noise), the PNSR are better compared to other non-AVE scenes (visible or audible).
For each backbone, the model with CDF and DNM showed the best performance.
I3D-CDF-DNM was the best of all the models.

The qualitative results of our I3D-CDF-DNM model is shown in Figure \ref{fig:non-ave}.
Our proposed model successfully suppresses the heatmap output in the two types of non-AVE frames.
In (a), the type of event changes from "speaking" to "playing flute" before and after the pause.
Our proposed model was able to follow the change and localize the correct sounding object.
In (c), at first, we hear the sound of a bus running but we don't see the bus, then the bus appears.
Our model was able to suppress the heatmap output while the bus is not visible.

\section{Discussion and Conclusion}
For the first time in this field, we created the annotations (AVSOL-E dataset) and the evaluation metrics for quantitative comparisons of AVSOL methods.
As the AVSOL-E dataset and evaluation source code will be publicly available, we hope it will be useful to other researchers.
The idea of sounding object presented in this paper is effective.
Although we have annotated 28 categories of events in AVE dataset this time, we think that datasets containing more types of events (e.g. VGGSound\cite{Chen20} with 310 categories) can also be annotated using the same idea.

Inspired by the findings from WSOL and previous audio-visual multimodal studies, we proposed DNM which uses MIL and attention at the same time to effectively connect AVC task and event classification task.
We evaluated two baseline methods and proposed models on AVSOL-E dataset.
Our models with DNM outperformed the baselines in all the evaluation metrics.
They also showed better performance compared to other methods and baselines in difficult scenes, \ie multiple sounding objects and three types of non-AVE segments.
Thanks to the new AVSOL-E dataset and the evaluation metrics, we were able to quantitatively and qualitatively demonstrate the superiority of our proposed method.

Our algorithm can be implemented to operate in real-time manner.
The ability to find sounding objects is a fundamental ability for living creatures.
The results of this research are expected to be applied to many applications, including real-world robots.

   {\small
      \bibliographystyle{ieee_fullname}
      \bibliography{egbib}
   }

\end{document}